\documentclass[review]{elsarticle}
\usepackage{graphicx}
\usepackage{booktabs}
\usepackage{subfig}
\usepackage{bm}
\usepackage{latexsym}
\usepackage{amssymb}
\usepackage{amsmath}
\usepackage{diagbox}
\usepackage[misc]{ifsym}
\usepackage{url}
\usepackage{xcolor}

\journal{Journal of \LaTeX\ Templates}

\begin{document}

\begin{frontmatter}

\title{Image Moment Invariants to Rotational Motion Blur}

\author[mymainaddress]{Hanlin Mo}
\author[mysecondaryaddress]{Hongxiang Hao}
\author[mymainaddress]{Guoying Zhao\corref{mycorrespondingauthor}}
\address[mymainaddress]{Center for Machine Vision and Signal Analysis, University of Oulu, Finland}
\address[mysecondaryaddress]{Key Laboratory of Intelligent Information Processing, Institute of Computing Technology, Chinese Academy of Sciences, Beijing, China}
\cortext[mycorrespondingauthor]{Corresponding author}
\ead{guoying.zhao@oulu.fi}

%

\begin{abstract}
Rotational motion blur caused by the circular motion of the camera or/and object is common in life. Identifying objects from images affected by rotational motion blur is challenging because this image degradation severely impacts image quality. Therefore, it is meaningful to develop image invariant features under rotational motion blur and then use them in practical tasks, such as object classification and template matching. This paper proposes a novel method to generate image moment invariants under general rotational motion blur and provides some instances. Further, we achieve their invariance to similarity transform. To the best of our knowledge, this is the first time that moment invariants for rotational motion blur have been proposed in the literature. We conduct extensive experiments on various image datasets disturbed by similarity transform and rotational motion blur to test these invariants' numerical stability and robustness to image noise. We also demonstrate their performance in image classification and handwritten digit recognition. Current state-of-the-art blur moment invariants and deep neural networks are chosen for comparison. Our results show that the moment invariants proposed in this paper significantly outperform other features in various tasks.     
\end{abstract}

\begin{keyword}
Blurred image \sep Rotational motion blur \sep Moment invariants \sep Spatial transform \sep Deep neural network \sep Image classification \sep Object recognition
\end{keyword}

\end{frontmatter}


\section{Introduction}
\label{section:1}
Feature extraction is one of the most challenging parts of image analysis. Different images of the same object can be captured using different cameras, from different viewpoints, and under different illumination conditions. These images may also be disturbed by additive noises and image blur. Ideal image features should be able to describe the object's intrinsic information, which means their numerical values should be invariant to image degradations caused by external factors. Researchers have defined numerous mathematical models to describe realistic image degradations in the past half-century and developed various invariant features under these models. Among these invariant features, moments and moment invariants play a crucial role.  

In a nutshell, moments are "projections" of an image function on a polynomial basis. For example, we can define geometric, complex, and orthogonal moments using the standard power basis, the polynomial basis of complex monomials, and orthogonal polynomial bases. Classical moment invariants are usually homogeneous polynomials of image moments and are invariant under specific image degradation. Much research has focused on constructing moment invariants under spatial transforms. Based on the theory of algebraic invariants, Hu derived seven geometric moment invariants of grayscale images under two-dimensional similarity transform (composed of rotation, scaling, and translation) \cite{1}. These invariants, known as Hu moments, have been widely used in various practical applications \cite{2,3,4,5}. The paper \cite{6} found it simpler to construct similarity moment invariants using complex moments. Additionally, several papers have proposed rotation or similarity invariants based on orthogonal moments, such as Zernike and Gaussian-Hermite moments \cite{7,8,9,10}. Reiss, Flusser, and Suk \cite{11,12} modified Hu's method and developed several affine moment invariants. They evaluated the performance of these invariants in planar objects and character recognition \cite{13,14}. Subsequently, researchers have devised more intuitive methods for generating affine moment invariants, such as the graph method \cite{15} and the geometric primitive method \cite{16}. Recently, Li et al. demonstrated the existence of projective moment invariants using finite combinations of image weighted moments \cite{17}. 

\begin{figure}
	\centering
	\includegraphics[height=25mm,width=120mm]{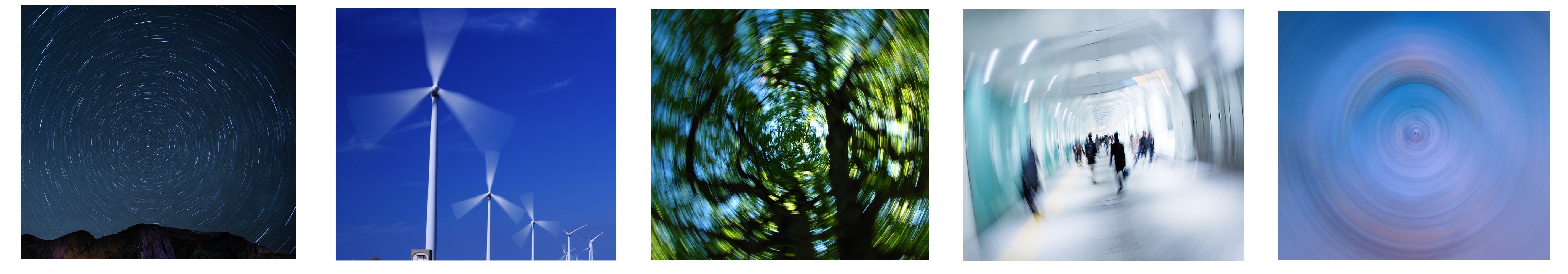}\\
	\caption{Some examples of rotational motion blur caused by the camera or object circular motion.}\label{figure:1}
\end{figure}

Unlike spatial transforms that alter the spatial coordinate system, some image degradations primarily affect the intensity values of an image, such as image noise, image blur, and color changes caused by illumination condition. Constructing moment invariants to image blur has received significant attention. Image blur can be broadly categorized as out-of-focus blur and motion blur. Out-of-focus blur is often caused by incorrect focus, a shallow depth of field, or a dirty lens and can be mathematically modeled by convolving a sharp image with a point spread function (PSF). Flusser et al. demonstrated how image geometric moments change under a convolutional operation \cite{18}. Assuming that the PSF $h(x,y)$ has central symmetry (i.e., $h(x,y)=h(-x,-y)$), they derived blur moment invariants that can be expressed as recursive functions of image geometric moments. These invariants have found applications in image registration \cite{19,20}, image forgery detection \cite{21}, and so on. In practice, image blurring often co-occurs with spatial transforms. The paper \cite{22} first developed complex moment invariants to both similarity transform and centrosymmetric blur. Subsequently, some researchers generated geometric and orthogonal moment invariants to both affine transform and centrosymmetric blur\cite{23,24,25,26}. Other papers supposed that the PSF is a uniform Gaussian function with circular symmetry and constructed moment invariants to both similarity/affine transform and Gaussian blur \cite{27,28,29,30}. Actually, a realistic PSF of out-of-focus blur often takes a form similar to a regular polygon, exhibiting $N$-fold rotational symmetry, which is determined by the aperture shape. In 2015, the paper \cite{31} proposed a complicated method for generating complex moment invariants to both similarity transform and $N$-fold rotational symmetric blur. For a $N$-fold symmetric PSF, Mo et al. proved that its geometric moments of the same order are linearly dependent \cite{32}. Based on this property, they further proposed a simple approach to determine if an existing similarity or affine geometric moment invariant also has invariance to $N$-fold symmetric blur.   

Compared to moment invariants for out-of-focus blur, there has been relatively little research on moment invariants for image blur caused by camera and/or object motion. Depending on the type of motion, image motion blur can be further divided into linear motion blur, rotational motion blur, radial motion blur, and others. Most previous studies concentrated on building image moment invariants to linear motion blur. In 1996, Flusser et al. derived geometric moment invariants to image blur induced by uniform linear motion \cite{33}. Similar to their previous work, this approach was also based on the assumption that uniform linear motion has central symmetry. These invariants have been employed in weed recognition, and wood slice recognition \cite{34,35}. Unlike linear motion blur, rotational motion blur results from the camera or object's rotation (circular motion) rather than its translation. It is commonly observed in daily life, and we show some examples in Figure \ref{figure:1}. Rotational motion blur can significantly affect the quality of the captured image, making it difficult to extract useful information for object recognition. Therefore, developing invariant features for rotational motion blur is meaningful. However, to our knowledge, no prior work has proposed image moment invariants under rotational motion blur. We aim to fill this research gap, and our contributions can be summarized as follows: 
\begin{itemize}
	\item We propose a novel method for constructing complex moment invariants to both similarity transform and rotational motion blur denoted as $RMBMIs$. Notably, we handle general rotational motion blur without imposing any restrictions on the properties of circular motion, such as requiring it to have a uniform rotational speed. 
	\item Using this method, we generate possible $RMBMIs$ up to the fourth/sixth order. Based on the relationship between complex moments and geometric moments, these $RMBMIs$ are further expressed as the functions of geometric moments. As a result, we first derive geometric moment invariants to both similarity transform and rotational motion blur.  
	\item We conduct extensive experiments on various image datasets disturbed by similarity transform and rotational motion blur to evaluate the stability of $RMBMIs$, their robustness to image noise, and their recognition ability in object classification and handwritten digit recognition. Our results show that $RMBMIs$ outperform current state-of-the-art blur moment invariants and deep neural networks in these tasks. 
\end{itemize}

The paper is organized as follows. Section \ref{section:2} provides some definitions and concepts for our work. Sections \ref{section:3}, \ref{section:4}, and \ref{section:5} are the main contribution of this paper. We develop a new method to generate complex and geometric moment invariants under both similarity transform and rotational motion blur. In Section \ref{section:6}, numerical experiments are conducted to validate our method further. Section \ref{section:7} concludes our work and discusses plans.  

\section{Basic concepts and definitions}
\label{section:2}
This section will introduce some basic concepts and definitions used in the following sections.
  
\subsection{Image Rotational Motion Blur} 
\label{section:2.1}
A grayscale image of an interested object can be regarded as a 2D scalar function $f(x,y):\Omega \subset \mathbb{R\times R} \rightarrow \mathbb{R}$. In the polar coordinate system, it can be expressed as $f(r,\theta)$, where $r=\sqrt{x^{2}+y^{2}}$ and $\theta= \mbox{arctan}\left(y/x\right)$. 

This paper analyzes the blurring caused by rotational motion around the object's center. Specifically, supposing that a sharp image $f(r,\theta)$ is disturbed by general rotational motion blur and $g(r,\theta)$ represents the blurred version, we have 
\begin{equation}\label{formula:1}
g(r,\theta)=\frac{1}{T}\int^{T}_{0}f(r,\theta-\psi(t))dt
\end{equation}
where $T$ represents the exposure time $(s)$ and $\psi(t)$ can be any function of $t$. 

For example, when the object (or the camera) rotates counterclockwise at a constant speed around its center, we have
\begin{equation}\label{formula:2}
\psi(t)=\omega t,~~~g(r,\theta)=\frac{1}{T}\int^{T}_{0}f(r,\theta-\omega t)dt
\end{equation}
where the constant $\omega$ represents the angular velocity $(rad/s)$. 

If the angular velocity $\omega$ is accelerated at a constant rate $\alpha$ $(rad/s^{2})$, we have 
\begin{equation}\label{formula:3}
\psi(t)=\omega t+\frac{1}{2}\alpha t^{2}~~~g(r,\theta)=\frac{1}{T}\int^{T}_{0}f(r,\theta-\omega t-\frac{1}{2}\alpha t^{2})dt  
\end{equation}

In many practical cases, the function expression $\psi(t)$ changes during the exposure time $T$. Thus, we can further extend (\ref{formula:1}) as follows 
\begin{equation}\label{formula:4}
g(r,\theta)=\frac{1}{T}\sum_{k=1}^{K}\int^{T_{k}}_{T_{k-1}}f(r,\theta-\psi_{k}(t))dt
\end{equation}
where $K$ is a positive integer, $0=T_{0}<T_{1}<T_{2}<\cdots<T_{K-1}<T_{K}=T$, and $\psi_{1},\psi_{2},\cdots,\psi_{K}$ are different functions in terms of $t$. When $K=1$, the formula (\ref{formula:4}) degenerates into (\ref{formula:1}). 

For example, supposing that the object (or the camera) rotates counterclockwise as defined by (\ref{formula:2}) when $0\leq t<\frac{T}{2}$ and rotates clockwise as defined by (\ref{formula:3}) when $\frac{T}{2}\leq t\leq T$, we have $\psi_{1}(t)=\omega t$ and $\psi_{2}(t)=-\omega(t-\frac{T}{2})-\frac{1}{2}\alpha (t-\frac{T}{2})^{2}$, meaning that
\begin{equation}\label{formula:5}
\begin{split}
g(r,\theta)&=\frac{1}{T}\left\{\int^{\frac{T}{2}}_{0}f(r,\theta-\omega t)dt+\right.\\&\left.\int^{T}_{\frac{T}{2}}f\left(r,\theta-\omega \frac{T}{2}+\omega \left(t-\frac{T}{2}\right)+\frac{1}{2}\alpha \left(t-\frac{T}{2}\right)^{2}\right)dt\right\}
\end{split}
\end{equation}       

\begin{figure}
	\centering
	\subfloat[Image blur caused by the uniform circular motion (\ref{formula:2}), when setting $(\omega, T)=(\frac{\pi}{20},1), (\frac{2\pi}{20},2), (\frac{3\pi}{20},3), (\frac{4\pi}{20},4), (\frac{5\pi}{20},5)$, respectively.]
	{\includegraphics[height=12.4mm,width=112mm]{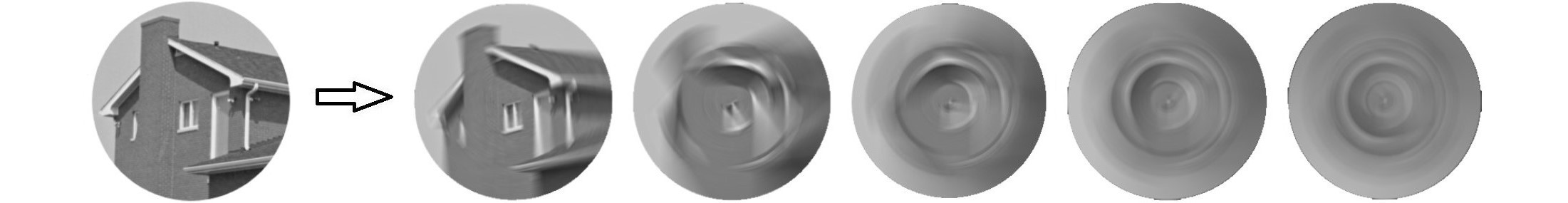}\label{figure:2(a)}\hfill}\\
	\subfloat[Image blur caused by the uniformly accelerated circular motion (\ref{formula:3}), when setting $(\omega, a, T)=(\frac{\pi}{20},\frac{\pi}{200},1), (\frac{2\pi}{20},\frac{2\pi}{200},2), (\frac{3\pi}{20},\frac{3\pi}{200},3), (\frac{4\pi}{20},\frac{4\pi}{200},4), (\frac{5\pi}{20},\frac{5\pi}{200},5)$, respectively.]
	{\includegraphics[height=12.4mm,width=112mm]{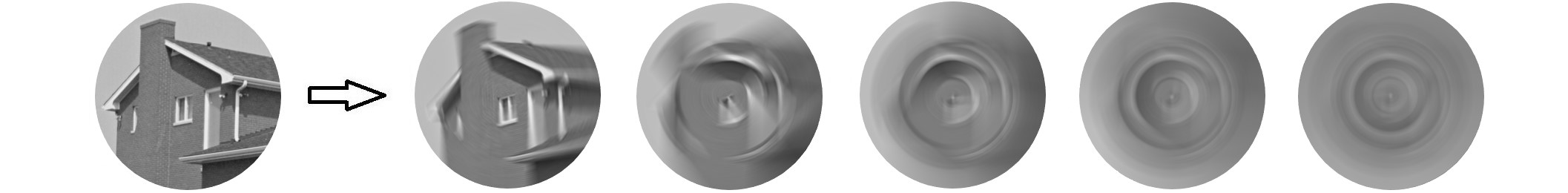}\label{figure:2(b)}\hfill}\\
	\subfloat[Image blur caused by the reciprocating circular motion (\ref{formula:5}). We set $(\omega, a, T)=(\frac{\pi}{20},\frac{\pi}{200},1), (\frac{2\pi}{20},\frac{2\pi}{200},2), (\frac{3\pi}{20},\frac{3\pi}{200},3), (\frac{4\pi}{20},\frac{4\pi}{200},4), (\frac{5\pi}{20},\frac{5\pi}{200},5)$, respectively.]
	{\includegraphics[height=12.5mm,width=110mm]{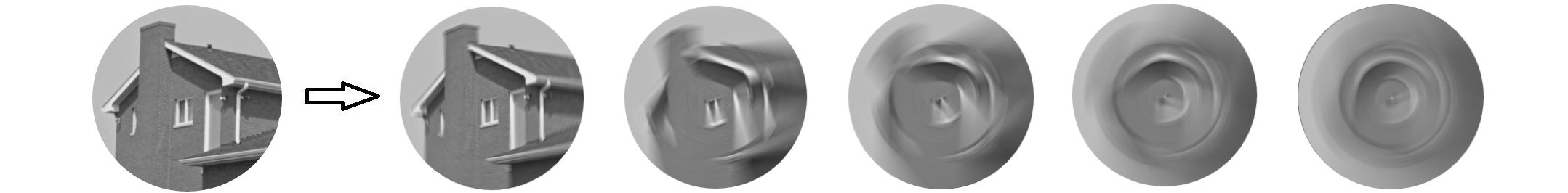}\label{figure:2(c)}\hfill}
	\caption{Some blurred images synthesized using the uniform circular motion (\ref{formula:2}), the uniformly accelerated circular motion (\ref{formula:3}), and the reciprocating circular motion (\ref{formula:5}).}\label{figure:2}
\end{figure}

In Figure \ref{figure:2}, we show various degraded versions of a sharp image by using (\ref{formula:2}), (\ref{formula:3}), and (\ref{formula:5}), respectively.         

\subsection{Image Moments}
\label{section:2.2}
This paper proposes a moment-based feature (i.e., moment invariants) that are invariant to rotational motion blur. Thus, it is necessary first to introduce the definition of image moments. As stated above, for an image function $f(x,y):\Omega \subset \mathbb{R\times R} \rightarrow \mathbb{R}$, its moments are "projections" of $f(x,y)$ on a polynomial basis. Specifically, let $\left\{\pi_{pq}(x,y)\right\}$ be a 2-variable polynomial basis of the space of $f(x,y)$ defined on $\Omega$, the general moments of $f(x,y)$ can be defined as 
\begin{equation}\label{formula:6}
	M^{(f)}_{pq}=\iint\limits_{\Omega}\pi_{pq}(x,y)f(x,y)dxdy
\end{equation}  
where $p$ and $q$ are non-negative integers and they show the highest power of $x$ and $y$ in $\pi_{pq}(x,y)$, respectively. Typically, the number $(p+q)$ is called the order of the moment. Note that we omit the superscript $(f)$ when there is no confusion.  

Commonly used image moments include geometric moments and complex moments. Image geometric moments $m_{pq}$ are defined based on the standard power basis $\pi_{pq}(x,y)=x^{p}y^{q}$
\begin{equation}\label{formula:7}
m_{pq}=\iint\limits_{\Omega}x^{p}y^{q}f(x,y)dxdy
\end{equation}    
And the polynomial basis of complex monomials $\pi_{pq}(x,y)=\left(x+iy\right)^{p}\left(x-iy\right)^{q}$ yields image complex moments $c_{pq}$
\begin{equation}\label{formula:8}
c_{pq}=\iint\limits_{\Omega}\left(x+iy\right)^{p}\left(x-iy\right)^{q}f(x,y)dxdy
\end{equation}
where $i$ represents the imaginary unit. Note that only the subscripts $p\geq q$ are independent and worth considering because $c_{pq}=c^{\star}_{qp}$ (the asterisk denotes complex conjugate). In the polar coordinate system, $c^{(f)}_{pq}$ can be expressed as  
\begin{equation}\label{formula:9}
c_{pq}=\int^{R}_{0}\int^{2\pi}_{0}r^{p+q+1}e^{i(p-q)\theta}f(r,\theta)d\theta dr
\end{equation}
where the positive real number $R$ represents the radius of the domain $\Omega$.

In theory, all polynomial bases are equivalent because they generate the same space of functions. This means that a moment with respect to a certain basis can be expressed in terms of moments with respect to any other basis \cite{36}. For example, image complex moments $c_{pq}$ and geometric moments $m_{pq}$ are related by
\begin{equation}\label{formula:10}
c_{pq}=\sum_{a=0}^{p}\sum_{b=0}^{q}\tbinom{p}{a}\tbinom{q}{b}(-1)^{q-b}\cdot i^{p+q-a-b}\cdot m_{a+b,p+q-a-b}
\end{equation}

In this paper, we first construct image moment invariants from complex moments to rotational motion blur and then derive the corresponding geometric moment invariants using the relationship (\ref{formula:10}).  

\subsection{Complex Moment Invariants to Image Similarity Transform}
\label{section:2.3}
As stated previously, moment invariants are special functions of image moments. They are invariant to certain image degradations, including spatial transforms, color changes, and image blur. In this paper, we first construct image moment invariants to rotational motion blur and then further prove their invariance to similarity transform. 

Two-dimensional similarity transform plays an essential role in commonly used image spatial transforms. It consists of two-dimensional rotation, scaling, and translation. In this paper, the origin of the spatial coordinate system is established at the center of the image, which has achieved translation invariance. As a result, we only need to consider two parameters, the rotation angle $\phi\in[0,2\pi)$ and the scale parameter $s\in\mathbb{R}^{+}$. Supposing that an image $f(r,\theta)$ defined on the polar coordinate system is transformed into $g(r^{'},\theta^{'})$ using a similarity transformation, we have $g(r^{'},\theta^{'})=f(\frac{r^{'}}{s},\theta^{'}-\phi)$.  

As far as we know, there has not been any previous research on image moment invariants to rotational motion blur, but much work about constructing similarity moment invariants. Earlier researchers constructed similarity moment invariants using normalized geometric moments $\widetilde{m}_{pq}=m_{pq}/\left(m^{(p+q)/2+1}_{00}\right)$, such as seven Hu moments \cite{1}. Flusser et al. found that constructing these invariants from normalized complex moments $\widetilde{c}_{pq}=c_{pq}/\left(c^{(p+q)/2+1}_{00}\right)$ is more straightforward and intuitive \cite{6}. In the polar coordinate system, they proved that $\widetilde{c}^{(g)}_{pq}=e^{i(p-q)\phi}\widetilde{c}^{(f)}_{pq}$. Let $n$ be a positive integer and $k_{i},p_{i},q_{i}$ be non-negative integers where $i=1,2,...,n$. When $\sum_{i=1}^{n}k_{i}(p_{i}-q_{i})=0$, the product of normalized complex moments $\prod^{n}_{i=1}\left(\widetilde{c}_{p_{i}q_{i}}/\widetilde{c}^{\left(p_{i}+q_{i}\right)/2+1}_{00}\right)^{k_{i}}$ is invariant to image similarity transform.   

\section{Complex Moment Invariants to Image Rotational Motion Blur}
\label{section:3}
First, let us show how a complex moment defined by (\ref{formula:9}) is changed under image rotational motion blur defined by (\ref{formula:4}).  
\\
\\
\textbf{Theorem 1.} Suppose that a sharp image $f(r,\theta)$ is disturbed by the rotational motion blur defined by (\ref{formula:4}) and $g(r,\theta)$ represents this blurred version.

Then, we have the following relationships: 
\begin{equation}\label{formula:11}
c^{(g)}_{pq}=
\left\{
\begin{array}{ll}
c^{(f)}_{pq}, & \mbox{~when~~}p=q,\\
\left(\frac{1}{T}\sum_{k=1}^{K}\int^{T_{k}}_{T_{k-1}}e^{i(p-q)\psi(t)}dt\right)c^{(f)}_{pq}, &\mbox{~when~~}p>q,
\end{array}
\right.  
\end{equation}
where the image complex moment $c_{pq}$ is defined by (\ref{formula:9}).  
\\
\\
\textbf{Proof:} According to (\ref{formula:4}) and (\ref{formula:9}), we have:
\begin{equation}\label{formula:12}
\begin{split}
c^{(g)}_{pq}&=\int^{R}_{0}\int^{2\pi}_{0}r^{p+q+1}e^{i(p-q)\theta}g(r,\theta)d\theta dr\\&
=\int^{R}_{0}\int^{2\pi}_{0}r^{p+q+1}e^{i(p-q)\theta}\left[\frac{1}{T}\sum_{k=1}^{K}\int^{T_{k}}_{T_{k-1}}f(r,\theta-\psi_{k}(t))dt\right] d\theta dr\\&
=\frac{1}{T}\sum_{k=1}^{K}\int^{T_{k}}_{T_{k-1}}\int^{R}_{0}\int^{2\pi}_{0}r^{p+q+1}e^{i(p-q)\theta}f(r,\theta-\psi_{k}(t)) d\theta dr dt\\&
=\frac{1}{T}\sum_{k=1}^{K}\int^{T_{k}}_{T_{k-1}}\int^{R}_{0}\int^{2\pi-\psi_{k}(t)}_{0-\psi_{k}(t)}r^{p+q+1}e^{i(p-q)(\theta+\psi_{k}(t))}f(r,\theta) d\theta dr dt\\&
=\frac{1}{T}\sum_{k=1}^{K}\int^{T_{k}}_{T_{k-1}}e^{i(p-q)\psi_{k}(t)}\left[\int^{R}_{0}\int^{2\pi-\psi_{k}(t)}_{0-\psi_{k}(t)}r^{p+q+1}e^{i(p-q)\theta}f(r,\theta) d\theta dr \right]dt\\&
=\frac{1}{T}\sum_{k=1}^{K}\int^{T_{k}}_{T_{k-1}}e^{i(p-q)\psi(t)}\left[\int^{R}_{0}\int^{2\pi}_{0}r^{p+q+1}e^{i(p-q)\theta}f(r,\theta) d\theta dr \right]dt\\&
=\left(\frac{1}{T}\sum_{k=1}^{K}\int^{T_{k}}_{T_{k-1}}e^{i(p-q)\psi_{k}(t)}dt\right)c^{(f)}_{pq}
\end{split}
\end{equation}
When $p=q$, we further have
\begin{equation}\label{formula:13}
c^{(g)}_{pp}=\left(\frac{1}{T}\sum_{k=1}^{K}\int^{T_{k}}_{T_{k-1}}1dt\right)c^{(f)}_{pp}=\left(\frac{1}{T}\sum_{k=1}^{K}\left(T_{k}-T_{k-1}\right)\right)c^{(f)}_{pp}=\frac{T}{T}c^{(f)}_{pp}=c^{(f)}_{pp}
\end{equation}	
The theorem is proved.
$\hfill\square$

Theorem 1 indicates that $c_{pp}$ is absolutely invariant to arbitrary rotational motion blur. When $p>q$, $c_{pq}$ just has relatively invariance. For some special functions $\psi_{k}(t)$, we can directly calculate the value of the definite integral $\int^{T_{k}}_{T_{k-1}}e^{i(p-q)\psi_{k}(t)}dt$. For example, when $K=1$ and $\psi_{1}(t)=\omega t$, we have
\begin{equation}\label{formula:14}
\frac{1}{T}\sum_{k=1}^{K}\int^{T_{k}}_{T_{k-1}}e^{i(p-q)\psi_{k}(t)}dt=\frac{1}{T}\int^{T}_{0}e^{i(p-q)\omega t}dt=\frac{i\left(T-e^{i(p-q)\omega T}\right)}{(p-q)\omega T^{2}}
\end{equation} 

To eliminate the constant $\left(\frac{1}{T}\sum_{k=1}^{K}\int^{T_{k}}_{T_{k-1}}e^{i(p-q)\psi_{k}(t)}dt\right)$ and obtain an absolute invariant, we can normalize a relative invariant by other relative invariants so that the constant get canceled. 
\\
\\
\textbf{Theorem 2.} Let a sharp image $f(r,\theta)$ be disturbed by the rotational motion blur defined by (\ref{formula:4}) and $g(r,\theta)$ represents this blurred version. Supposing that $p_{i},q_{i},p_{j},q_{j}$ are different non-negative integers and $p_{i}-q_{i}=p_{j}-q_{j}>0$, we have   
\begin{equation}\label{formula:15}
\frac{c^{(g)}_{p_{i}q_{i}}}{c^{(g)}_{p_{j}q_{j}}}=\frac{c^{(f)}_{p_{i}q_{i}}}{c^{(f)}_{p_{j}q_{j}}}
\end{equation} 
where the image complex moment $c_{pq}$ is defined by (\ref{formula:9}).
\\
\\
\textbf{Proof:} According to Theomrem 1, when $p_{i}-q_{i}=p_{j}-q_{j}>0$, we have:
\begin{equation}\label{formula:16}
\frac{c^{(g)}_{p_{i}q_{i}}}{c^{(g)}_{p_{j}q_{j}}}=\frac{\left(\frac{1}{T}\sum_{k=1}^{K}\int^{T_{k}}_{T_{k-1}}e^{i(p_{i}-q_{i})\psi_{k}(t)}dt\right)c^{(f)}_{p_{i}q_{i}}}{\left(\frac{1}{T}\sum_{k=1}^{K}\int^{T_{k}}_{T_{k-1}}e^{i(p_{j}-q_{j})\psi_{k}(t)}dt\right)c^{(f)}_{p_{j}q_{j}}}=\frac{c^{(f)}_{p_{i}q_{i}}}{c^{(f)}_{p_{j}q_{j}}}
\end{equation}	
The theorem is proved.
$\hfill\square$

Using Theorems 1 and 2, we can generate all possible complex moment invariants $c_{pp}$ and $\frac{c_{p_{i}q_{i}}}{c_{p_{j}q_{j}}}$ up to a given order. For example, when setting the order $(p+q)\leq 4$, we derive a set of five complex moment invariants to rotational motion blur as follows
\begin{equation}\label{formula:17}
RMBMI^{4}=\left\{c_{00}, c_{11},c_{22}, \frac{c_{10}}{c_{21}}, \frac{c_{20}}{c_{31}}\right\}
\end{equation}
Similarly, when setting $(p+q)\leq 6$, we have
\begin{equation}\label{formula:18}
RMBMI^{6}=\left\{c_{00}, c_{11}, c_{22}, c_{33}, \frac{c_{10}}{c_{21}}, \frac{c_{20}}{c_{31}}, \frac{c_{10}}{c_{32}}, \frac{c_{30}}{c_{41}},\frac{c_{20}}{c_{42}}, \frac{c_{40}}{c_{51}}\right\}
\end{equation} 

It should be noted that the product or sum of these complex moment invariants are also invariant to rotational motion blur, such as $(c_{11}+c_{22})$ and $\frac{c_{10}c_{20}}{c_{21}c_{31}}$. However, there are different types of dependencies between these more complicated invariants and the simple ones that make up them, including linear, polynomial, and functional dependencies. Hence, in this paper, we only generate and analyze the most fundamental invariants $c_{pp}$ and $\frac{c_{p_{i}q_{i}}}{c_{p_{j}q_{j}}}$.   

\section{Complex Moment Invariants to Both Similarity Transform and Rotational Motion Blur}
\label{section:4}
In many practical applications, such as image classification, object recognition, and template matching, we have to handle various types of image degradations simultaneously. For example, in some cases, one sharp image and one blurred image of the same object are taken from different spatial positions. To extract intrinsic information from these two images, we need to construct moment invariants that are invariant simultaneously to certain spatial transforms and image blur. As Section \ref{section:2.3} mentions, image similarity transform is a commonly used geometric transform model in computer vision and pattern recognition. Thus, in this section, we further analyze the properties of complex moments $c_{pq}$ under both similarity transform and rotational motion blur. 
\\
\\
\textbf{Theorem 3.} Suppose that a sharp image $f(r,\theta)$ is first transformed using a similarity transformation and then further disturbed by the rotational motion blur defined by (\ref{formula:4}). Let $g(r^{'},\theta^{'})$ represent this degraded version ($r^{'}\in[0,R^{'}]$ and $\theta^{'}\in[0,2\pi)$), meaning that 
\begin{equation}\label{formula:19}
g(r^{'},\theta^{'})=\frac{1}{T}\sum_{k=1}^{K}\int^{T_{k}}_{T_{k-1}}f\left(\frac{r^{'}}{s},\theta^{'}-\phi-\psi_{k}(t)\right)dt
\end{equation}
where $\phi$ and $s$ represent the rotation angle and scale parameter in the similarity transformation, respectively. 

Then, we have the following relationships
\begin{equation}\label{formula:20}
	\widetilde{c}^{(g)}_{pq}=
	\left\{
	\begin{array}{ll}
		\widetilde{c}^{(f)}_{pq}, & \mbox{~when~~}p=q,\\
		\left(\frac{e^{i(p-q)\phi}}{T}\sum_{k=1}^{K}\int^{T_{k}}_{T_{k-1}}e^{i(p-q)\psi_{k}(t)}dt\right) \widetilde{c}^{(f)}_{pq}, &\mbox{~when~~}p>q,
	\end{array}
	\right.  
\end{equation}
where the normalized complex moment $\widetilde{c}_{pq}$ is defined as
\begin{equation}\label{formula:21}
\widetilde{c}_{pq}=\frac{c_{pq}}{\left(c^{\left(p+q\right)/2+1}_{00}\right)}
\end{equation}
and the complex moment $c_{pq}$ is defined by (\ref{formula:9}).
\\
\\
\textbf{Proof:} According to (\ref{formula:9}) and (\ref{formula:19}), we have
\begin{equation}\label{formula:22}
\begin{split}
c^{(g)}_{pq}&=\int^{R^{'}}_{0}\int^{2\pi}_{0}(r^{'})^{p+q+1}e^{i(p-q)\theta^{'}}g(r^{'},\theta^{'})d\theta^{'} dr^{'}\\&
=\int^{R^{'}}_{0}\int^{2\pi}_{0}(r^{'})^{p+q+1}e^{i(p-q)\theta^{'}}\left[\frac{1}{T}\sum_{k=1}^{K}\int^{T_{k}}_{T_{k}}f(\frac{r^{'}}{s},\theta^{'}-\phi-\psi_{k}(t))dt\right] d\theta^{'} dr^{'}\\&
=\frac{1}{T}\sum_{k=1}^{K}\int^{T_{k}}_{T_{k-1}}\int^{R^{'}}_{0}\int^{2\pi}_{0}(r^{'})^{p+q+1}e^{i(p-q)\theta^{'}}f(\frac{r^{'}}{s},\theta^{'}-\phi-\psi_{k}(t)) d\theta^{'} dr^{'} dt\\&
=\frac{1}{T}\sum_{k=1}^{K}\int^{T_{k}}_{T_{k-1}}\int^{R}_{0}\int^{2\pi-\phi-\psi_{k}(t)}_{0-\phi-\psi_{k}(t)}(s\cdot r)^{p+q+1}e^{i(p-q)(\theta+\phi+\psi_{k}(t))}f(r,\theta) d\theta d(s\cdot r) dt\\&
=\frac{s^{p+q+2}e^{i(p-q)\phi}}{T}\sum_{k=1}^{K}\int^{T_{k}}_{T_{k-1}}e^{i(p-q)\psi_{k}(t)}\left[\int^{R}_{0}\int^{2\pi-\phi-\psi_{k}(t)}_{0-\phi-\psi_{k}(t)}r^{p+q+1}e^{i(p-q)\theta}f(r,\theta) d\theta dr \right]dt\\&
=\frac{s^{p+q+2}e^{i(p-q)\phi}}{T}\sum_{k=1}^{K}\int^{T_{k}}_{T_{k-1}}e^{i(p-q)\psi_{k}(t)}\left[\int^{R}_{0}\int^{2\pi}_{0}r^{p+q+1}e^{i(p-q)\theta}f(r,\theta) d\theta dr \right]dt\\&
=s^{p+q+2}\cdot \left(\frac{e^{i(p-q)\phi}}{T}\sum_{k=1}^{K}\int^{T_{k}}_{T_{k-1}}e^{i(p-q)\psi_{k}(t)}dt\right)c^{(f)}_{pq}
\end{split}
\end{equation}	
Thus, the normalized complex moment $\widetilde{c}^{(g)}_{pq}$ satisfies the following relationship
\begin{equation}\label{formula:23}
\begin{split}
\widetilde{c}^{(g)}_{pq}&=\frac{c^{(g)}_{pq}}{\left(c^{(g)}_{00}\right)^{\left(p+q\right)/2+1}}\\&
=\frac{s^{p+q+2}\cdot \left(\frac{e^{i(p-q)\phi}}{T}\sum_{k=1}^{K}\int^{T_{k}}_{T_{k-1}}e^{i(p-q)\psi_{k}(t)}dt\right)c^{(f)}_{pq}}{\left(s^{2}\cdot \left(\frac{1}{T}\sum_{k=1}^{K}\int^{T_{k}}_{T_{k-1}}1dt\right)c^{(f)}_{00}\right)^{(p+q)/2+1}}\\&
=\frac{s^{p+q+2}\cdot \left(\frac{e^{i(p-q)\phi}}{T}\sum_{k=1}^{K}\int^{T_{k}}_{T_{k-1}}e^{i(p-q)\psi_{k}(t)}dt\right)c^{(f)}_{pq}}{\left(s^{2}\cdot \left(\frac{1}{T}\sum_{k=1}^{K}\left(T_{k}-T_{k-1}\right)\right)c^{(f)}_{00}\right)^{(p+q)/2+1}}\\&
=\frac{s^{p+q+2}\cdot \left(\frac{e^{i(p-q)\phi}}{T}\sum_{k=1}^{K}\int^{T_{k}}_{T_{k-1}}e^{i(p-q)\psi_{k}(t)}dt\right)c^{(f)}_{pq}}{s^{p+q+2}\cdot\left(c^{(f)}_{00}\right)^{(p+q)/2+1}}\\&
=\left(\frac{e^{i(p-q)\phi}}{T}\sum_{k=1}^{K}\int^{T_{k}}_{T_{k-1}}e^{i(p-q)\psi_{k}(t)}dt\right)\frac{c^{(f)}_{pq}}{\left(c^{(f)}_{00}\right)^{(p+q)/2+1}}\\&
=\left(\frac{e^{i(p-q)\phi}}{T}\sum_{k=1}^{K}\int^{T_{k}}_{T_{k-1}}e^{i(p-q)\psi_{k}(t)}dt\right)\widetilde{c}^{(f)}_{pq}
\end{split}
\end{equation}
When $p=q$, we further have 
\begin{equation}\label{formula:24}
\widetilde{c}^{(g)}_{pp}=\left(\frac{1}{T}\sum_{k=1}^{K}\int^{T_{k}}_{T_{k-1}}1dt\right)\widetilde{c}^{(f)}_{pp}=\left(\frac{1}{T}\sum_{k=1}^{K}\left(T_{k}-T_{k-1}\right)\right)\widetilde{c}^{(f)}_{pp}=\frac{T}{T}\widetilde{c}^{(f)}_{pp}=\widetilde{c}^{(f)}_{pp}
\end{equation}	
The theorem is proved.
$\hfill\square$

Theorem 3 shows that $\widetilde{c}_{pp}$ is invariant simultaneously to similarity transform and rotational motion blur. When $p>q$, we can also use the approach in Theorem 2 to eliminate the constant $\left(\frac{e^{i(p-q)\phi}}{T}\sum_{k=1}^{K}\int^{T_{k}}_{T_{k-1}}e^{i(p-q)\psi_{k}(t)}dt\right)$.   
\\
\\
\textbf{Theorem 4.} Let a sharp image $f(r,\theta)$ be first transformed using a similarity transform and then be disturbed by the rotational motion blur defined by (\ref{formula:4}) while $g(r^{'},\theta^{'})$ represents this degraded version. 

Supposing that $p_{i},q_{i},p_{j},q_{j}$ are different non-negative integers and $p_{i}-q_{i}=p_{j}-q_{j}>0$, we have   
\begin{equation}\label{formula:25}
\frac{\widetilde{c}^{(g)}_{p_{i}q_{i}}}{\widetilde{c}^{(g)}_{p_{j}q_{j}}}=\frac{\widetilde{c}^{(f)}_{p_{i}q_{i}}}{\widetilde{c}^{(f)}_{p_{j}q_{j}}}
\end{equation} 
where $\widetilde{c}_{pq}$ is defined by (\ref{formula:21}).
\\
\\
\textbf{Proof:} According to Theorem 3, when $p_{i}-q_{i}=p_{j}-q_{j}>0$, we have:
\begin{equation}\label{formula:26}
\frac{\widetilde{c}^{(g)}_{p_{i}q_{i}}}{\widetilde{c}^{(g)}_{p_{j}q_{j}}}=\frac{\left(\frac{e^{i(p_{i}-q_{i})\phi}}{T}\sum_{k=1}^{K}\int^{T_{k}}_{T_{k-1}}e^{i(p_{i}-q_{i})\psi_{k}(t)}dt\right)\widetilde{c}^{(f)}_{p_{i}q_{i}}}{\left(\frac{e^{i(p_{j}-q_{j})\phi}}{T}\sum_{k=1}^{K}\int^{T_{k}}_{T_{k-1}}e^{i(p_{j}-q_{j})\psi_{k}(t)}dt\right)\widetilde{c}^{(f)}_{p_{j}q_{j}}}=\frac{\widetilde{c}^{(f)}_{p_{i}q_{i}}}{\widetilde{c}^{(f)}_{p_{j}q_{j}}}
\end{equation}	
The theorem is proved.
$\hfill\square$

In summary, Theorems 3 and 4 illustrate that all complex moment invariants $\widetilde{c}_{pp}$ and $\frac{\widetilde{c}_{p_{i}q_{i}}}{\widetilde{c}_{p_{j}q_{j}}}$ are invariant simultaneously to similarity transform and rotational motion blur. For each invariant listed in the sets $RMBMI^{4}$ and $RMBMI^{6}$ defined by (\ref{formula:17}) and (\ref{formula:18}), we can directly replace $c_{pq}$ with $\widetilde{c}_{pq}$ to further achieve its invariance under similarity transform. In this case, $c_{00}$ can no longer be used as an independent invariant because it has been used to normalize the other complex moments $c_{pq}$. As far as we know, it is the first time that this type of moment invariant has been proposed in the literature. 

\section{Geometric Moment Invariants to Both Similarity Transform and Rotational Motion Blur}
\label{section:5}
As stated in Section \ref{section:2.2}, the most commonly used image moments are geometric and complex moments. In Sections \ref{section:3} and \ref{section:4}, we have proposed a simple method to construct moment invariants from complex moments under rotational motion blur and to achieve their invariance to similarity transform further. In contrast, it is not easy to directly build the corresponding geometric moment invariants. Previous research has found that each complex moment can be expressed in geometric moments as (\ref{formula:10}). Thus, a feasible approach is to replace each complex moment that a complex moment invariant depends upon with the corresponding expansion of geometric moments. 

We use this approach to handle five complex moment invariants in the set $RMBMI_{4}$ defined by (\ref{formula:17}) and finally obtain the following seven geometric moment invariants $RMBMI_{0}\sim RMBMI_{6}$ to rotational motion blur. Each complex moment invariant $\frac{c_{p_{i}q_{i}}}{c_{p_{j}q_{j}}}$ is a complex number containing real and imaginary parts. Hence, two geometric moment invariants can be obtained from each $\frac{c_{p_{i}q_{i}}}{c_{p_{j}q_{j}}}$.
\begin{equation}\label{formula:27}
\begin{split}
&RMBMI_{0}=c_{00}=m_{00}\\&
RMBMI_{1}=c_{11}=m_{20}+m_{02}\\&
RMBMI_{2}=c_{22}=m_{40}+2m_{22}+m_{04}\\&
RMBMI_{3}=Re\left(\frac{c_{10}}{c_{21}}\right)=\frac{m_{10}(m_{30}+m_{12})+m_{01}(m_{21}+m_{03})}{(m_{30}+m_{12})^{2}+(m_{21}+m_{03})^{2}}\\&
RMBMI_{4}=Im\left(\frac{c_{10}}{c_{21}}\right)=\frac{m_{10}(m_{21}+m_{03})-m_{01}(m_{30}+m_{12})}{(m_{30}+m_{12})^{2}+(m_{21}+m_{03})^{2}}\\&
RMBMI_{5}=Re\left(\frac{c_{20}}{c_{31}}\right)=\frac{(m_{20}-m_{02})(m_{40}-m_{04})+4m_{11}(m_{31}+m_{13})}{(m_{40}-m_{04})^2+4(m_{31}+m_{13})^2}\\&
RMBMI_{6}=Im\left(\frac{c_{20}}{c_{31}}\right)=\frac{2m_{11}(m_{40}-m_{04})-2(m_{20}-m_{02})(m_{31}+m_{13})}{(m_{40}-m_{04})^2+4(m_{31}+m_{13})^2}
\end{split}
\end{equation}	
To achieve the invariance of $RMBMI_{1}\sim RMBMI_{6}$ to similarity transform, we can also replace $m_{pq}$ with the corresponding normalized geometric moment $\widetilde{m}_{pq}=m_{pq}/(m_{00})^{(p+q)/2+1}$. 

The order of geometric moments these invariants depend upon is less than or equal to $4$. As mentioned above, Hu moments are the most famous image moment-based features widely used in many practical tasks. They are seven geometric moment invariants up to the third order and are invariant to similarity transform. Some of them appear in (\ref{formula:27}). For example, $RMBMI_{1}$ is the second Hu moment, and the denominator of $RMBMI_{3}$ and $RMBMI_{4}$ is the fourth Hu moment. In a similar fashion, we also derive $16$ geometric moment invariants (including $m_{00}$) up to the sixth order from the set $RMBMI^{6}$ defined by (\ref{formula:18}).             

\section{Experiment and Discussion}
\label{section:6}
In this section, we conduct extensive experiments on various blurred image datasets to verify the numerical stability of $RMBMIs$ and their robustness to image noise. We also evaluate their performance in flower image classification and handwritten digit recognition. State-of-the-art moment-based features and deep neural networks are chosen for comparison.

\subsection{The Stability and Robustness of RMBMIs}
\label{section:6.1}
\begin{figure}
	\centering
	\includegraphics[height=11mm,width=120mm]{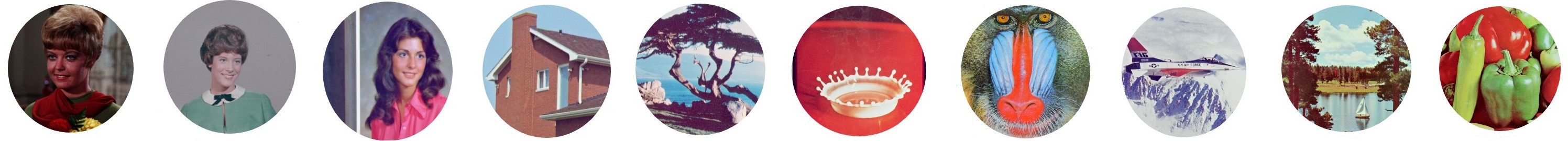}\\
	\caption{Ten original images selected from the USC-SIPI image database.}\label{figure:3}
\end{figure}

In previous sections, we always suppose that an image $f(x,y)$ is a continuous function and prove Theorems 1 to 4 based on this assumption. However, in practical tasks, $RMBMIs$ are calculated from digital images defined on a discrete domain. It is necessary to test if they still have good invariance in this case through numerical experiments. As shown in Figure \ref{figure:3}, we randomly select ten color images from the USC-SIPI image dataset (\url{http://sipi.usc.edu/database/}) and convert and resize them to grayscale images of $257\times 257$ pixels. Using the uniform circular motion (UCM) defined by (\ref{formula:2}), we generate fifty blurred versions for each original image by setting $\left(\omega,T\right)\in \{\frac{\pi}{20},\frac{2\pi}{20},\cdots,\frac{9\pi}{20},\frac{10\pi}{20}\}\times \{1,2,3,4,5\}$. Some blurred examples have been shown in Figure \ref{figure:2(a)}. 

Six geometric moment invariants $RMBMI_{1}\sim RMBMI_{6}$ defined by (\ref{formula:27}) are calculated from each original image and blurred image. Note that unless otherwise specified, we always use normalized geometric moments $\widetilde{m}_{pq}$ to calculate $RMBMIs$ in the following sections, which means they are also invariant to similarity transform. Then, we utilize the mean relative error ($MRE$) to quantify the numerical stability of each $RMBMI_{k}$. For any $k\in\{1,2,\cdots,6\}$, $MRE_{k}$ is defined as
\begin{equation}\label{formula:28}
	MRE_{k}=\frac{1}{10}\sum_{i=1}^{10}\frac{1}{50}\left(\sum_{j=1}^{50}\frac{\left|RMBMI_{k}(Img_{i})-RMBMI_{k}(Img^{j}_{i})\right|}{\left|RMBMI_{k}(Img_{i})\right|+\left|RMBMI_{k}(Img^{j}_{i})\right|}\right)\times 100\%
\end{equation}      
Here, $Img_{i}$ represents the $i$-th original image and $Img^{j}_{i}$ represents a blurred version of $Img_{i}$, where $i=1,2,\cdots,10$, and $j=1,2,\cdots,50$. Obviously, we have $0\leq MRE_{k}\leq 1$, and smaller $MRE_{k}$ indicates that $RMBMI_{k}$ has a better stability. 

Also, we need to test the invariance of $RMBMI_{k}$ to both similarity transform and rotational motion blur. To this end, we first randomly generate $50$ similarity transformations by setting the rotation angle $\theta\in\{\frac{\pi}{6},\frac{2\pi}{6},\cdots,\frac{9\pi}{6},\frac{10\pi}{6}\}$ and the scaling parameter $s\in\{0.6,0.8,1.0,1.2,1.4\}$. Then, we use these similar transformations to generate 50 transformed versions for each image. These transformed versions are further disturbed by UCM with the same parameter settings as before. Finally, we calculate $MRE_{k}$ again.  

The numerical values of six $MRE_{1}\sim MRE_{6}$ in these two cases are listed in the first column of Table \ref{table:1} (N: without similarity transform; Y: with similarity transform). First, we can see that all $RMBMI_{k}$ have excellent invariance to UCM because the maximum value of $MRE_{k}$ is only $2.274\%$ (when $k=6$). Secondly, $MRE_{1}$ and $MRE_{2}$ are less than $0.002\%$, while $MRE_{3}\sim MRE_{6}$ are between $0.123\% \sim 2.274\%$. This means that the numerical stability of $RMBMI_{1}$ and $RMBMI_{2}$ is better than $RMBMI_{3}\sim RMBMI_{6}$. Note that the first two invariants have more simple expressions or depend upon lower-order geometric moments than the latter four invariants. Thirdly, when images are disturbed by both similarity transform and rotation motion blur, all $MRE_{k}$ increase obviously but are still less than $3.673\%$. The extra errors are caused by the interpolation operation when rotating and scaling images. 

Besides UCM, Section \ref{section:2.1} also introduces another two types of rotational motion blur, the uniformly accelerated circular motion (UACM) defined by (\ref{formula:3}) and the reciprocating circular motion (RCM) defined by (\ref{formula:5}). We use UACM and RCM to generate blurred images by setting $(\omega,a,T)=\left\{\frac{\pi}{20},\frac{2\pi}{20},\cdots,\frac{9\pi}{20},\frac{10\pi}{20}\right\}\times\left\{\frac{\pi}{200},\frac{2\pi}{200},\cdots,\frac{9\pi}{200},\frac{10\pi}{200}\right\}\times\left\{1,2,3,4,5\right\}$, and test the stability of six invariants under them again. Some blurred examples caused by these two motion models are shown in Figures \ref{figure:2(b)} and \ref{figure:2(c)}, respectively. The results are listed in the second and the third columns of Table \ref{table:1}. We can see that they are consistent with the results under UCM. Thus, $RMBMIs$ maintain good numerical stability when calculated on discrete images.

\begin{table}
	\caption{\label{table:1} The numerical stability of $RMBMI_{1}\sim RMBMI_{6}$ under three types of rotational motion blur (N: without similarity transform; Y: with similarity transform).}
	\centering
	\begin{tabular}{lcccccc}
		\toprule[1.3pt]
		  & \multicolumn{2}{c}{\textbf{UCM~~(\ref{formula:2})}} &  \multicolumn{2}{c}{\textbf{UACM~~(\ref{formula:3})}} &  \multicolumn{2}{c}{\textbf{RCM~~(\ref{formula:5})}}\\
		\textbf{MRE} & \textbf{N} & \textbf{Y}  & \textbf{N} & \textbf{Y} & \textbf{N} & \textbf{Y}\\
		\toprule[1.3pt]
		\bm{$MRE_{1}$} & 0.001\% & 0.007\% & 0.001\% & 0.007\% & 0.001\% & 0.007\%\\
		\bm{$MRE_{2}$} & 0.002\% & 0.015\% & 0.002\% & 0.015\% & 0.002\% & 0.015\%\\
		\bm{$MRE_{3}$} & 0.123\% & 0.264\% & 0.043\% & 0.193\% & 0.028\% & 0.191\%\\
		\bm{$MRE_{4}$} & 1.465\% & 3.673\% & 0.818\% & 3.260\% & 0.801\% & 3.249\%\\
		\bm{$MRE_{5}$} & 0.470\% & 0.583\% & 0.148\% & 0.306\% & 0.101\% & 0.281\%\\
		\bm{$MRE_{6}$} & 2.274\% & 3.026\% & 0.926\% & 2.066\% & 0.922\% & 2.014\%\\
		\midrule
	\end{tabular}
\end{table}

\begin{figure}
	\centering
	\includegraphics[height=33mm,width=120mm]{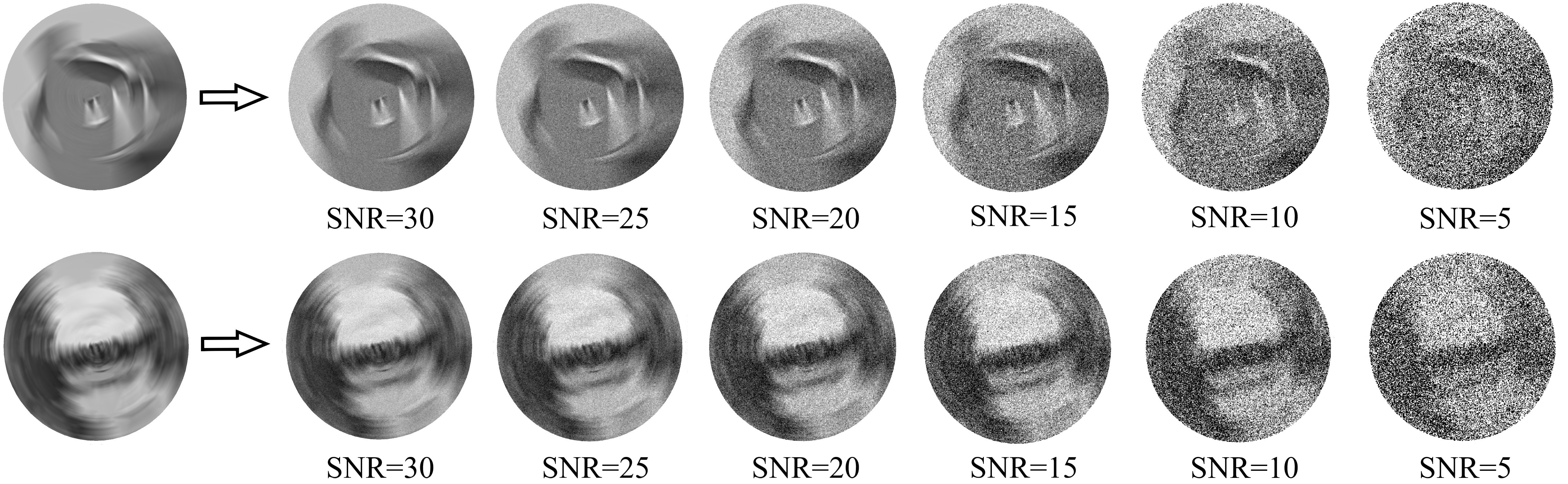}\\
	\caption{Each $Img^{j}_{i}$ is corrupted by different levels of Gaussian white noise.}\label{figure:4}
\end{figure}

\begin{figure}
	\centering
	\subfloat[UCM+Gaussian white noise.]
	{\includegraphics[height=42mm,width=57mm]{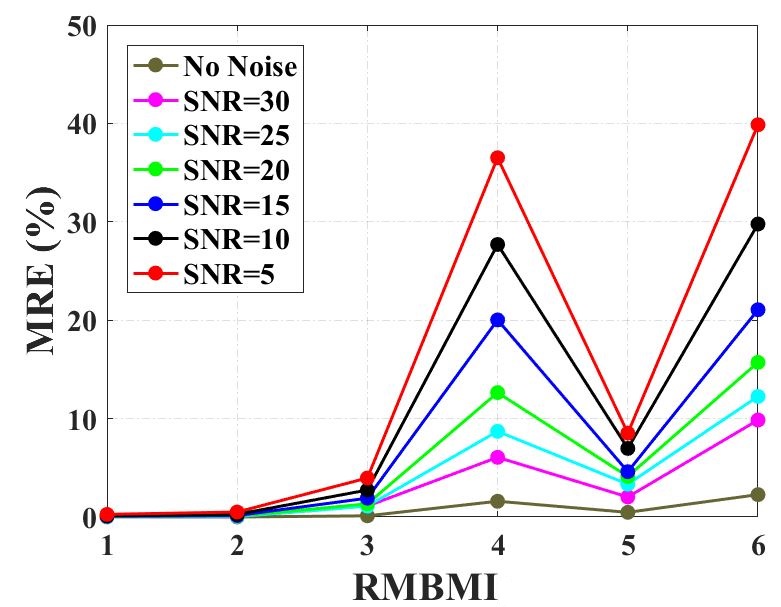}\label{figure:5(a)}\hfill}~~~~
	\subfloat[Similarity transform+UCM+Gaussian white noise.]
	{\includegraphics[height=42mm,width=57mm]{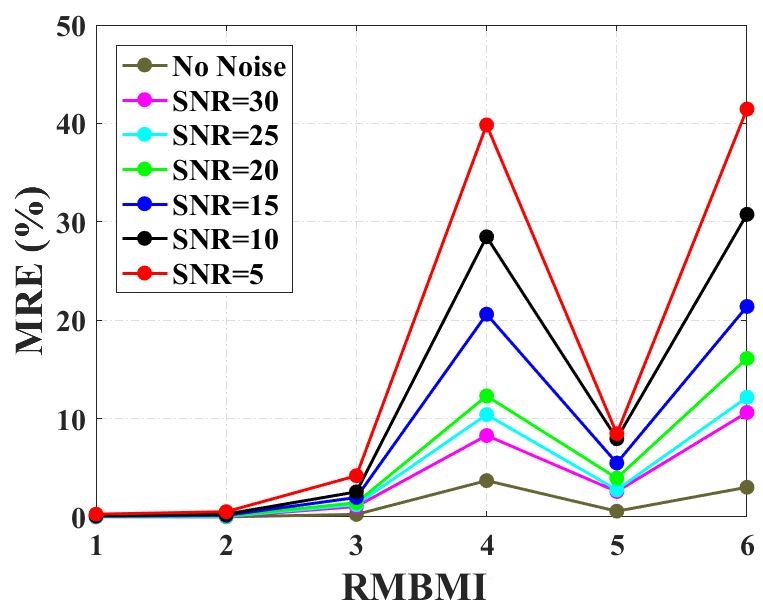}\label{figure:5(b)}\hfill}
	\caption{By adding different levels of Gaussian white noise to each $Img^{j}_{i}$ disturbed by UCM (or both similar transform and UCM), we calculate $MRE_{1}\sim MRE_{6}$ again to quantify the robustness of $RMBMI_{1}\sim RMBMI_{6}$ to image noise.}\label{figure:5}
\end{figure}

Image noise is a random variation in the brightness or color of an image, caused by poor lighting conditions, long exposure times, and other factors. It can degrade image quality and negatively impact the performance of image features. Like previous studies \cite{38,39,40}, we also test the robustness of $RMBMIs$ to image noise. As shown in Figure \ref{figure:4}, we add different levels of Gaussian white noise ($SNR=30,25,20,15,10,5$) to each $Img^{j}_{i}$ disturbed by UCM (or both similarity transform and UCM), and then recalculate $MRE_{1}\sim MRE_{6}$. The results are shown in Figures \ref{figure:5(a)} and \ref{figure:5(b)}. Firstly, we found that $RMBMI_{1}$ and $RMBMI_{2}$ have strong robustness to noise. Even when $Img^{j}_{i}$ is heavily disturbed by Gaussian white noise, such as when $SNR=5$, the values of $MRE_{1}$ and $MRE_{2}$ are still less than $0.3\%$. Secondly, as $SNR$ increases, $MRE_{3}\sim MRE_{6}$ keep increasing, but $MRE_{3}$ and $MRE_{5}$ are always less than $10\%$, while the performance of $MRE_{4}$ and $MRE_{6}$ is much worse, reaching up to nearly $40\%$. This indicates that although $RMBMI_{3}$ and $RMBMI_{4}$, $RMBMI_{5}$ and $RMBMI_{6}$ are constructed using the same geometric moments, the robustness of $RMBMI_{3}$ and $RMBMI_{5}$ is significantly better than that of $RMBMI_{4}$ and $RMBMI_{6}$.  

\begin{figure}
	\centering
	\subfloat[Twenty training images selected from the Oxford Flower dataset.]
	{\includegraphics[height=21mm,width=120mm]{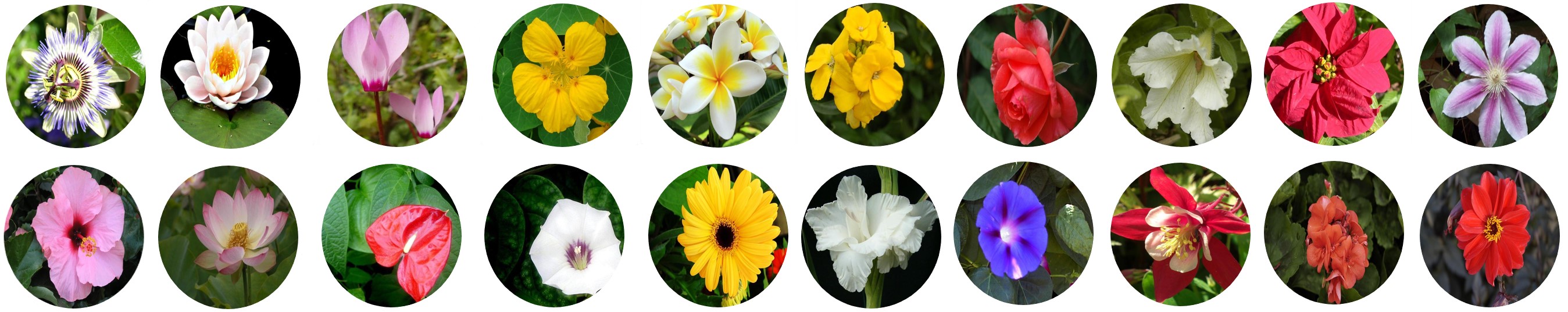}\label{figure:6(a)}\hfill}\\
	\subfloat[Test images disturbed by both similarity transform and UCM.]
	{\includegraphics[height=13mm,width=120mm]{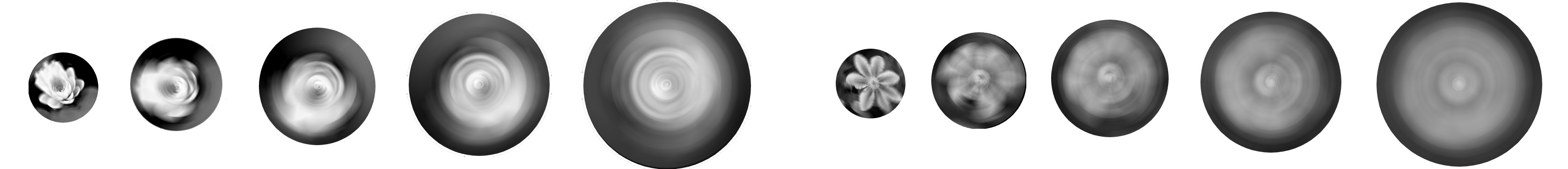}\label{figure:6(b)}\hfill}\\
	\subfloat[Test images disturbed by both similarity transform and UACM.]
	{\includegraphics[height=13mm,width=120mm]{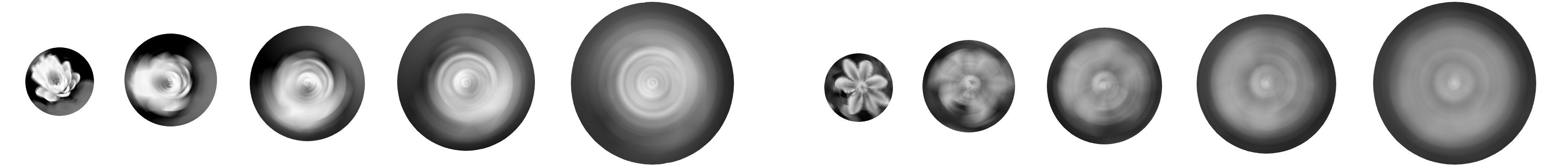}\label{figure:6(c)}\hfill}\\
	\subfloat[Test images disturbed by both similarity transform and RCM.]
	{\includegraphics[height=13mm,width=120mm]{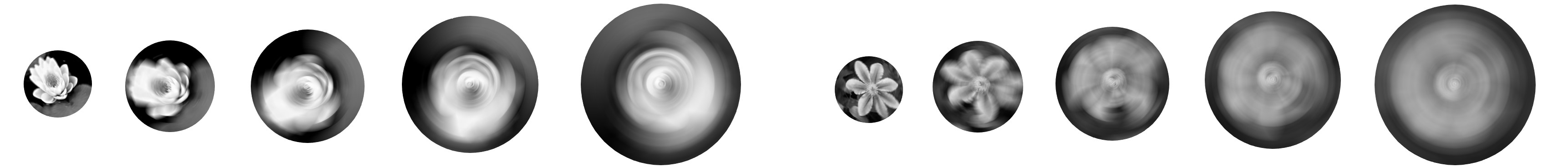}\label{figure:6(d)}\hfill}\\
	\caption{Some training and test images used for image classification.}\label{figure:6}
\end{figure}

\subsection{The Recognition Ability of RMBMIs}
\label{section:6.2}
This subsection evaluates the recognition ability of $RMBMIs$ in image classification and compares their performance with other types of blur moment invariants. We conduct the classification experiment on the Oxford Flower dataset  (\url{https://www.robots.ox.ac.uk/~vgg/data/flowers/}). This image database contains $102$ different categories of flowers. We randomly select one image from each of the top twenty categories as our training images, a total of 20 images (see Figure \ref{figure:6(a)}). Each is converted to a grayscale image and resized to $257\times 257$ pixels. Then, using the same parameter settings in Section \ref{section:6.1}, we generate $50$ blurred versions for each training image by using UCM, UACM, and RCM, respectively. This process
yields three test datasets, each containing $20\times50 = 1000$ test images. Also, we can first transform each test image using a random similarity transformation and then disturb it by UCM, UACM, or RCM. Figures \ref{figure:6(b)}, \ref{figure:6(c)}, and \ref{figure:6(d)} show several degraded images in these three test datasets, respectively. 

We calculate $RMBMI_{1}\sim RMBMI_{6}$ on each training and test image, which are used as a feature vector, and then utilize the nearest neighbor classifier for image classification. Following the previous papers \cite{17,32}, the modified Chi-square distance is used to measure the similarity of training and test images in the space of features. Besides $RMBMIs$, current state-of-the-art blur moment invariants are chosen for comparison. To our knowledge, no previous literature has proposed moment invariants under rotational motion blur. Hence, we select six types of moment invariants under centrosymmetric blur, circularly symmetric blur, $N$-fold rotational symmetric blur, or linear motion blur. Some of them also have invariance to similarity or affine transform.  
\begin{itemize}
	\item $BMs$ (9 dimensions): Nine blur moments $\big(B(1,3),B(3,1),B(4,0),B(3,2),\\B(2,3),B(4,1),B(1,4),B(5,0),B(0,5)\big)$ proposed in \cite{18}, which are invariant to centrosymmetric blur.
	\item $LMIs$ (10 dimensions): The first ten Legendre moment invariants $\big(\overline{I}(3,0),\\\overline{I}(2,1),\overline{I}(1,2),\overline{I}(0,3)
	,\overline{I}(5,0),\overline{I}(4,1),\overline{I}(3,2),\overline{I}(2,3),\overline{I}(1,4),\overline{I}(0,5)\big)$ proposed in \cite{37}, which are invariant to centrosymmetric blur.
	\item $CBAMIs$ (5 dimensions): Five combined invariants $\big(I_{1},I_{2},I_{3},I_{4},I_{5}\big)$ proposed in \cite{24}, which have invariance to both affine transform and centrosymmetric blur.
	\item $CMIs$ (6 dimensions): Six complex moments $\left(\widetilde{c}_{20},\widetilde{c}_{02},\widetilde{c}_{30},\widetilde{c}_{21},\widetilde{c}_{12},\widetilde{c}_{03}\right)$ proposed in \cite{31} which are invariant to both similarity transform and $N$-fold rotational symmetric blur ($N>3$).
	\item $HMs^{5}$ (5 dimensions): The second to sixth Hu moments proposed in \cite{1}. Our previous work \cite{32} proved that they are invariant simultaneously to similarity transform and $N$-fold rotational symmetric blur ($N>3$).
	\item $LMBMIs$ (4 dimensions): The second, third, fifth and seventh Hu moments. The paper \cite{33} proved that they are invariant to both similarity transform and linear motion blur. 
\end{itemize}

\begin{table}
	\caption{\label{table:2} The classification accuracies from various blur moment invariants on the UCM, UACM and RCM test sets (N: without similarity transform; Y: with similarity transform). Bold and underline stand for best and second-best results.}
	\centering
	\begin{tabular}{lcccccc}
		\toprule[1.3pt]
		& \multicolumn{2}{c}{\textbf{UCM~~(\ref{formula:2})}} &  \multicolumn{2}{c}{\textbf{UACM~~(\ref{formula:3})}} &  \multicolumn{2}{c}{\textbf{RCM~~(\ref{formula:5})}}\\
		\textbf{Feature} & \textbf{N} & \textbf{Y}  & \textbf{N} & \textbf{Y} & \textbf{N} & \textbf{Y}\\
		\toprule[1.3pt]
		\bm{$BMs$} \cite{18} & 23.60\% & 7.40\% & 22.70\% & 7.80\% & 45.50\% & 9.30\%\\
		\bm{$LMIs$} \cite{37} & 25.80\% & 7.40\% & 24.50\% & 6.40\% & 47.70\% & 10.00\%\\
		\bm{$CBAMIs$} \cite{24} & 35.10\% & 34.90\% & 31.80\% & 31.70\% & 57.90\% & 58.10\%\\
		\bm{$CMIs$} \cite{31} & \underline{39.30\%} & \underline{38.60\%} & \underline{39.70\%} & 36.00\% & 63.40\% & 63.40\%\\
		\bm{$HMs^{5}$} \cite{32} & \underline{39.30\%} & 36.00\% & 36.10\% & \underline{36.90\%} & \underline{64.60\%} & \underline{64.20\%}\\
		\bm{$LMBMIs$} \cite{33} & 34.70\% & 34.40\% & 30.90\% & 30.60\% & 59.00\% & 58.80\%\\
		\midrule
		\bm{$RMBMIs$} & \textbf{100\%} & \textbf{100\%} & \textbf{100\%} & \textbf{100\%} & \textbf{100\%} & \textbf{100\%}\\
		\midrule
	\end{tabular}
\end{table}

The classification accuracy rates from different moment-based features are summarized in Table \ref{table:2}. Firstly, it can be seen that six $RMBMIs$ achieve $100\%$ classification accuracy and significantly outperform the other types of blur moment invariants in every case, which is consistent with our theoretical analysis. Secondly, all blur moment invariants used for comparison achieve their own best results on the RCM test set $(45\sim 65\%)$, while performing poorly on the UCM and UACM test sets $(20\sim 40\%)$. From Figures \ref{figure:6(b)}, \ref{figure:6(c)}, and \ref{figure:6(d)}, it can be observed that RCM has a milder impact on image quality compared to UCM and UACM. In fact, when the acceleration $\alpha$ is small, the formula (\ref{formula:5}) is approximately equal to $2/T\int^{T/2}_{0}f(r,\theta-\omega t)dt$. It can be seen that, in this case, the exposure time is actually reduced from $T$ to $T/2$. Thirdly, $BMs$ and $LMIs$ are almost out of order when test images are deformed under random similarity transformations because they are only invariant to image blur. This underscores the significance of constructing moment invariants which are invariant to both spatial deformations and image blur. 

Subsequently, each test image is disturbed by different levels of Gaussian white noise ($SNR = 30, 25, 20, 15, 10, 5$), and we re-evaluate the performance of $RMBMIs$. In the previous section, we have found that $RMBMI_{4}$ and $RMBMI_{6}$ have poor noise robustness, so here we only use $RMBMI_{1}$, $RMBMI_{2}$, $RMBMI_{3}$ and $RMBMI_{5}$ as image features. As shown in Table $\ref{table:3}$, even with $SNR = 5$, the classification accuracy from the four $RMBMIs$ is still greater than $80\%$, particularly achieving a classification accuracy greater than $92\%$ on two RCM test datasets (with or without similarity transform). This once again demonstrates the robustness of $RMBMIs$ to image noise. 

\begin{table}
	\caption{\label{table:3} The classification accuracies from $RMBMIs$ on the UCM, UACM and RCM test sets disturbed by different levels of Gaussian white noise (N: without similarity transform; Y: with similarity transform).}
	\centering
	\begin{tabular}{p{10mm}<{\centering}p{14mm}<{\centering}p{14mm}<{\centering}p{14mm}<{\centering}p{14mm}<{\centering}p{14mm}<{\centering}p{14mm}<{\centering}}
		\toprule[1.3pt]
		& \multicolumn{2}{c}{\textbf{UCM~~(\ref{formula:2})}} &  \multicolumn{2}{c}{\textbf{UACM~~(\ref{formula:3})}} &  \multicolumn{2}{c}{\textbf{RCM~~(\ref{formula:5})}}\\
		\textbf{SNR} & \textbf{N} & \textbf{Y}  & \textbf{N} & \textbf{Y} & \textbf{N} & \textbf{Y}\\
		\toprule[1.3pt]
		\bm{$30$}\textbf{dB} & 96.30\% & 96.60\% & 99.30\% & 99.10\% & 99.80\% & 99.70\%\\
		\bm{$25$}\textbf{dB} & 95.00\% & 94.60\% & 98.90\% & 98.20\% & 99.60\% & 99.90\%\\
		\bm{$20$}\textbf{dB} & 93.80\% & 95.50\% & 97.20\% & 96.20\% & 99.20\% & 99.40\%\\
		\bm{$15$}\textbf{dB} & 93.50\% & 92.70\% & 94.50\% & 95.50\% & 97.80\% & 98.10\%\\
		\bm{$10$}\textbf{dB} & 89.10\% & 89.40\% & 89.20\% & 90.50\% & 95.80\% & 95.50\%\\
		\bm{$5$}\textbf{dB} & 83.90\% & 82.60\% & 82.50\% & 83.40\% & 92.40\% & 92.70\%\\
		\midrule
	\end{tabular}
\end{table}

\begin{figure}
	\centering
	\subfloat[Some examples in the training set.]
	{\includegraphics[height=30mm,width=55mm]{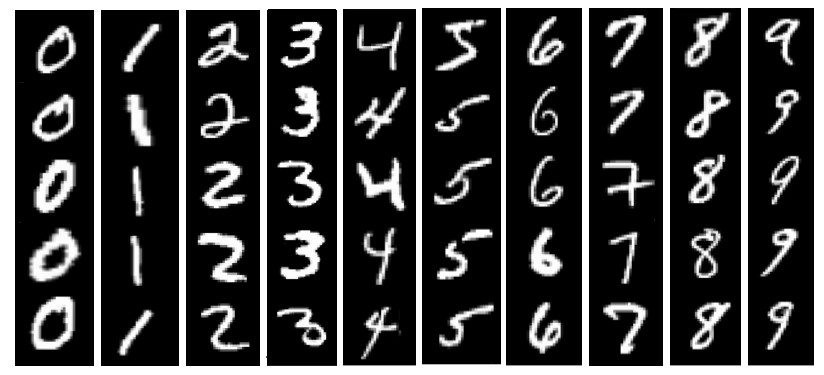}\label{figure:7(a)}\hfill}~~~~
	\subfloat[Some examples in the test set.]
	{\includegraphics[height=30mm,width=55mm]{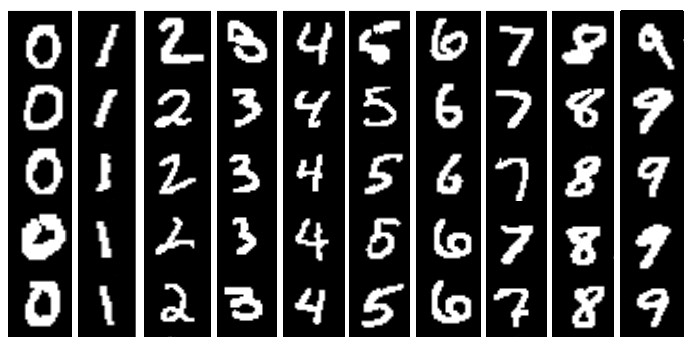}\label{figure:7(b)}\hfill}
	\caption{The training set and test set of the MNIST dataset.}\label{figure:7}
\end{figure}

\subsection{The Performance of RMBMIs in Handwritten Digit Recognition}
\label{section:6.3}
\begin{figure}
	\centering
	\includegraphics[height=40mm,width=115mm]{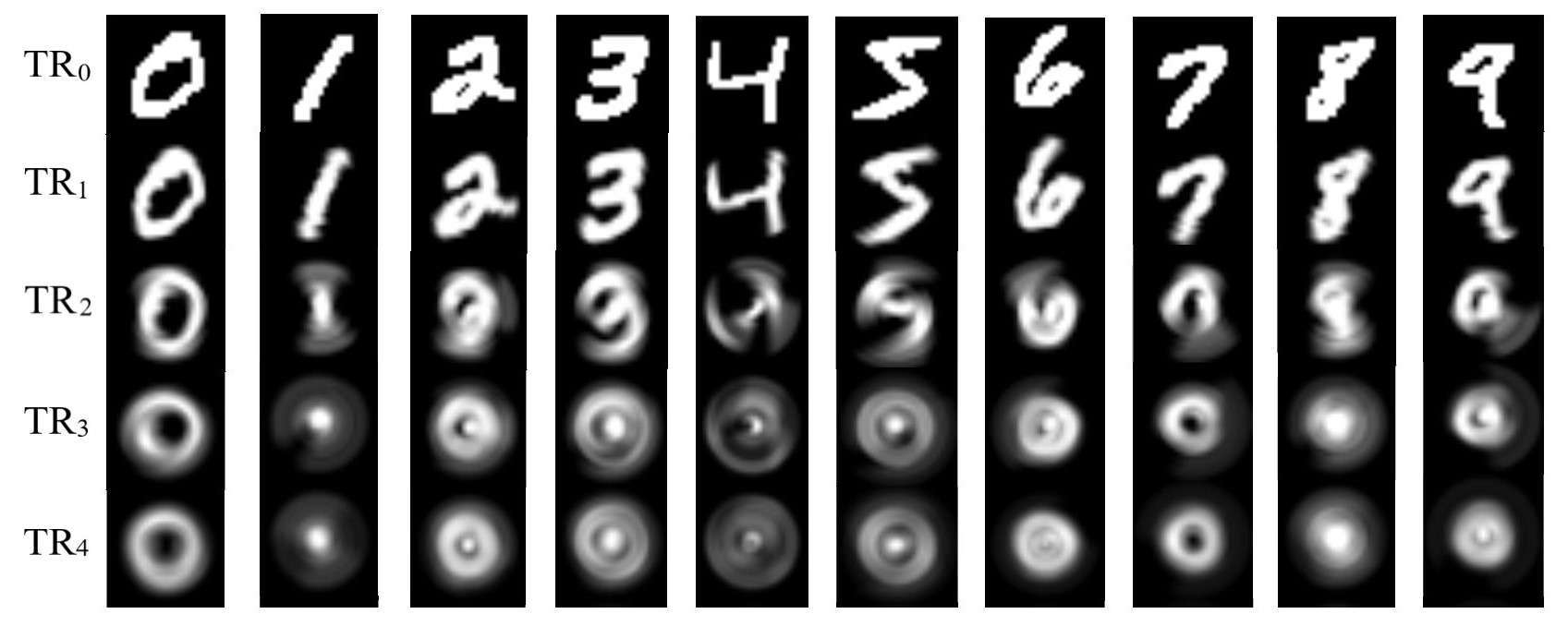}\\
	\caption{Some samples from different training sets $TR_{0}\sim TR_{4}$.}\label{figure:8}
\end{figure}

Since 2012, deep neural networks, especially convolutional neural networks, have been successfully applied in many pattern recognition tasks and significantly outperform hand-crafted image features. Thus, in the last experiment, we conduct handwritten digit classification based on the MNIST dataset \cite{41} and compare the performance of $RMBMIs$ with classical convolutional neural networks. This dataset consists of $70000$ grayscale images of handwritten digits from $0$ to $9$, with $60000$ training images and $10000$ test images. The size of these images is $28\times28$ pixels. Some examples are shown in Figures \ref{figure:7(a)} and \ref{figure:7(b)}. In this subsection, the original training and test sets are referred to as $TR_{0}$ and $TS_{0}$, respectively. To test the invariance of different methods to rotational motion blur, we further apply UCM defined by (\ref{formula:2}) to each training and test image. We set four different pairs of parameters for UCM, i.e., $(\omega,T)=(\frac{\pi}{10},1),(\frac{2\pi}{10},2),(\frac{3\pi}{10},3),(\frac{4\pi}{10},4)$, and obtain four new training sets $TR_{1}\sim TR_{4}$, and four new test sets $TS_{1}\sim TS_{4}$. Figure \ref{figure:8} shows some samples from all five training sets $TR_{0}\sim TR_{4}$.  

Compared to the datasets used in Sections \ref{section:6.1} and \ref{section:6.2}, the size of the MNIST dataset is relatively large. Hence, we calculate $RMBMIs$ up to the sixth order for each training and test image. As described in Section \ref{section:5}, we obtain 16 geometric moment invariants from the set $RMBMI^{6}$ defined by (\ref{formula:18}). Except for $m_{00}$, the remaining 15 invariants composed of normalized geometric moments $\widetilde{m}_{pq}$ are used as an image feature vector. Then, this 15-dimensional feature vector is input to a fully connected neural network (RMBMI-FCNN). It consists of four fully connected layers with $128$, $128$, $64$, and $64$ units, respectively. Batch normalization and ReLU activation functions are added after each fully connected layer, and the output layer contains $10$ units for classification. During training and testing, 15 $RMBMIs$ of each image are subtracted by their means and divided by their standard deviations across the entire training set. For comparison, we also design a standard convolutional neural network (CNN) that takes images as input, which consists of six convolutional layers with $32$, $32$, $64$, $64$, $128$, and $128$ kernels, respectively, with a size of $3\times3$. A $2\times 2$ max pooling is added after the second and fourth convolutional layers, and an $8\times8$ average pooling is added after the sixth layer. Finally, there is a fully connected layer with ten units. We also use batch normalization and ReLU activation functions after each convolutional layer. RMBMI-FCNN and CNN contain $3.4$K and $288.6$K learnable parameters, respectively.     
  
\begin{table}
	\caption{\label{table:4} The classification accuracies from RMBMI-FCNN on various traning sets $TR_{i}$ and test sets $TS_{j}$, where $i,j=0,1,\cdots,4$.}
	\centering
	\begin{tabular}{p{12mm}<{\centering}p{16mm}<{\centering}p{16mm}<{\centering}p{16mm}<{\centering}p{16mm}<{\centering}p{16mm}<{\centering}}
		\toprule[1.3pt]
		 & \textbf{TS$_{0}$} & \textbf{TS$_{1}$}  & \textbf{TS$_{2}$} & \textbf{TS$_{3}$} & \textbf{TS$_{4}$}\\
		\toprule[1.3pt]
		\textbf{TR$_{0}$} & 86.18\% & 85.80\% & 85.81\% & 85.38\% & 84.28\%\\
		\textbf{TR$_{1}$} & 86.81\% & 87.05\% & 86.80\% & 86.21\% & 85.30\%\\
		\textbf{TR$_{2}$} & 85.40\% & 85.16\% & 85.53\% & 85.31\% & 84.01\%\\
		\textbf{TR$_{3}$} & 85.97\% & 85.94\% & 86.35\% & 86.63\% & 84.25\%\\
		\textbf{TR$_{4}$} & 85.53\% & 85.32\% & 85.11\% & 84.94\% & 85.93\% \\
		\midrule
	\end{tabular}
\end{table}

We use the same protocol to train RMBMI-FCNN and CNN on specific $TR_{i}$, where $i=0,1,\cdots,4$, and then test their performance on all five $TS_{0}\sim TS_{4}$. Specifically, the cross-entropy loss is used, and the number of epochs and the batch size are $200$. The Adam optimizer is selected for optimization while the initial learning rate is $10^{-3}$, multiplied by $0.5$ every $40$ epochs. The classification accuracies from RMBMI-FCNN and CNN on various training and test sets are summarized in Tables \ref{table:4} and \ref{table:5}, respectively. When $i=j$, i.e., training and test sets are disturbed by the same UCM (with the same parameter setting), the CNN achieves high classification accuracies $97.30\%\sim99.40\%$, while the accuracies of the RMBMI-FCNN are $85.53\%\sim87.05\%$. The main reasons for this phenomenon are \textbf{(1)} through data augmentation, CNN can learn the invariance to a certain level of rotation motion blur; \textbf{(2)} 15 $RMBMIs$ up to the sixth order cannot extract all information contained in an image. In theory, we could use higher-order $RMBMIs$, but previous research has found their numerical stability poor. \textbf{(3)} As shown in Figures \ref{figure:7(a)} and \ref{figure:7(b)}, there are complex deformations between digits of the same class in the training and test sets, and similarity transform cannot well model such realistic deformations. 

\begin{table}
	\caption{\label{table:5} The classification accuracies from CNN on various traning sets $TR_{i}$ and test sets $TS_{j}$, where $i,j=0,1,\cdots,4$.}
	\centering
	\begin{tabular}{p{12mm}<{\centering}p{16mm}<{\centering}p{16mm}<{\centering}p{16mm}<{\centering}p{16mm}<{\centering}p{16mm}<{\centering}}
		\toprule[1.3pt]
		& \textbf{TS$_{0}$} & \textbf{TS$_{1}$}  & \textbf{TS$_{2}$} & \textbf{TS$_{3}$} & \textbf{TS$_{4}$}\\
		\toprule[1.3pt]
		\textbf{TR$_{0}$} & 99.24\% & 98.27\% & 50.11\% & 27.18\% & 21.52\%\\
		\textbf{TR$_{1}$} & 88.50\% & 99.40\% & 74.58\% & 18.12\% & 13.97\%\\
		\textbf{TR$_{2}$} & 19.96\% & 43.08\% & 99.02\% & 44.00\% & 29.46\%\\
		\textbf{TR$_{3}$} & 10.31\% & 19.68\% & 38.01\% & 97.67\% & 61.57\%\\
		\textbf{TR$_{4}$} & 12.87\% & 16.61\% & 29.10\% & 59.20\% & 97.30\% \\
		\midrule
	\end{tabular}
\end{table}

However, when $i\neq j$, i.e., training and test sets are blurred with different levels of UCM, the performance of the CNN drops drastically. For example, when the CNN is trained on $TR_{2}$, its accuracies on $TS_{0}$, $TS_{1}$, $TS_{3}$, and $TS_{4}$ are $19.96\%$, $43.08\%$, $44.00\%$, and $29.46\%$, respectively. In contrast, the performance of the RMBMI-FCNN hardly changed and remained between $84\%$ and $87\%$. This indicates that CNN's invariance to UCM depends entirely on training data. Obviously, in order to make it invariant to any level of UCM and, further, to any rotational motion blur, we need to augment the training set with all possible rotational motions and parameter settings, which would require a huge amount of computational resources to train the network. In addition, since the numerical values of parameters such as angular velocity, angular acceleration, and exposure time can be arbitrary in theory, we cannot generate all possible blurred images. On the other hand, the invariance of the RMBMI-FCNN does not depend on training data because the input features are naturally invariant to general rotational motion blur. This is one of the advantages of hand-crafted invariant features over deep neural networks.

\section{Conclusions}
\label{section:7}
This paper presents a novel method to construct complex and geometric moment invariants under image rotational motion blur ($RMBMIs$), which fills a research gap in the field of moments and moment invariants, and generates some instances of $RMBMIs$ up to the fourth/sixth order. Further, we achieve the invariance of $RMBMIs$ to similarity transform. Our experiments verify the numerical stability of $RMBMIs$ on discrete images and test their robustness to image noise. The results show that $RMBMIs$ outperform current state-of-the-art blur moment invariants and deep neural networks in object classification and handwritten digit recognition tasks. In the future, we plan to validate the performance of $RMBMIs$ on real-world blur images, integrate these invariants with deep neural networks better, and apply them to more practical tasks in computer vision.  

\section*{Acknowledgments}
This work has partly been funded by the National Key R\&D Program of China (No. 2017YFB1002703), the National Natural Science Foundation of China (Grant No. 60873164, 61227802 and 61379082) and the Academy of Finland for Academy Professor project EmotionAI (Grant No. 336116).

\bibliography{mybibfile}

\end{document}